# Bilingual Words and Phrase Mappings for Marathi and Hindi SMT

**Sreelekha S, Pushpak Bhattacharyya**

Indian Institute of Technology (IIT) Bombay, India

{sreelekha, pb}@cse.iitb.ac.in

**Abstract**

Lack of proper linguistic resources is the major challenges faced by the Machine Translation system developments when dealing with the resource poor languages. In this paper, we describe effective ways to utilize the lexical resources to improve the quality of statistical machine translation. Our research on the usage of lexical resources mainly focused on two ways, such as; augmenting the parallel corpus with more vocabulary and to provide various word forms. We have augmented the training corpus with various lexical resources such as lexical words, function words, kridanta pairs and verb phrases. We have described the case studies, evaluations and detailed error analysis for both Marathi to Hindi and Hindi to Marathi machine translation systems. From the evaluations we observed that, there is an incremental growth in the quality of machine translation as the usage of various lexical resources increases. Moreover, usage of various lexical resources helps to improve the coverage and quality of machine translation where limited parallel corpus is available.

## 1. Introduction

When translating text or speech from one natural language to another, how well the grammatical structures and linguistic properties are handled will determine the quality of the Machine Translation (MT). Complexity of the morphological structures and the linguistic divergence creates difficulties in MT (Kunchukuttan et.al., 2012; Ramananthan et. al., 2011; Dorr et. al., 1994; Och et. al., 2001; Ooch et. al., 2003; Knight K. 1999). Lack of proper linguistic resources to cover the various linguistic phenomena is the major problem faced by Indian language MT in addition to the rich morphology. Since India has 18 constitutional languages, which are written in 10 different scripts (Antony P. J. 2013), there is a large demand of MT system developments between Indian –Indian languages as well as from English to Indian languages (Nair, et.al., 2012). There are many ongoing attempts to develop MT systems for various Indian languages using rule-based as well as statistical-based approaches (Antony P. J. 2013; Ashan et. al., 2010; Brown et. al., 1993; Sreelekha et. al., 2013; Sreelekha et. al., 2015; Sreelekha et. al., 2015; Sreelekha et. al., 2016; Sreelekha et. al., 2017; Sreelekha et. al., 2018). This paper discusses various approaches used in Indian language to Indian language MT systems especially in Marathi to Hindi Statistical MT system and vice versa to improve the quality of machine translation. Marathi is morphologically very complex with agglutinative suffixes compared to Hindi. It has more than 200 nominal inflections and 450 verbal inflections which will cover about 10000 nouns and 1900 verbs in addition to the 175 postpositions attached to the nominal and verbal entities(Dixit et. al., 2005).
Consider the translated Hindi output from Marathi-Hindi SMT system for the Marathi sentence,

*Marathi* - तो घरी जाणाऱ्यांबरोबर जाई.
{to ghari jaanaryaabarobar jaie}
{he also used to go with the ones who used to go home}

*Hindi*- वह घर |UNK| जाता ।
{yah ghar jaatha vah naraz ho}

Here the Marathi word "जाणाऱ्यांबरोबर" {*jaanaryaabarobar*} {*go with the ones who used to go*}is failed to translate to its correct Hindi translation "जाने वालो के साथ" {*jaane walo ke saath*}{*go with the ones who used to go*} and is shown as unknown word. Also the verb जाई *{hota} is* wrongly mapped to "जाता"{*jaatha*} instead of "जाता था"{*jaatha tha*}. In this kind of situations lexical resources can play a major role to learn the various inflected forms. If we are able to train the machine with the verb phrase translation of "जाणाऱ्यांबरोबर" {*jaanaryaabarobar*} as "जाने वालो के साथ" {*jaane walo ke saath*} then it will help the machine to learn the inflections correctly.

## 2. Scenario of Statistical MT Research

In the case of Statistical Machine Translation, statistical models works purely based upon the frequency of occurrence and the alignments between source and target language words (Brown et. al., 1993; Kunchukuttan et.al., 2012; Ramananthan et. al., 2011; Dorr et. al., 1994; Och et. al., 2001; Ooch et. al., 2003; Knight K. 1999). Even though Indian languages follow same SOV order, there are many structural and vocabulary differences between languages. Moreover there are challenges of ambiguities such as; Lexical ambiguity, Structural ambiguity and Semantic ambiguity. In the case of Marathi- Hindi language pair, Marathi have a feature of dropping of subordinate clause called Participial Constructions, which means, constructions in Hindi having Participials in Marathi. Lexical resources can be helpful to handle this kind of difficulties and make the machine to learn different morphological word forms. Various categories of word forms such as lexical words, verb phrases, semantic words, morphological forms etc can be used. Importance of adding dictionary words to the corpus and its effect in improving the translation quality was studied by Och and Ney in their paper (Och and Ney, 2003). We have used IndoWordnet (Bhattacharyya 2010) semantic linked words as dictionary words to make the system learn various word forms. We have explained the extraction of various lexical resources and its augmentation process in machine translation in the experimental Section. The comparative performance analysis with phrase based model with that of augmented lexical resources is described in Section 3 & 4.

## 3. Experimental Discussion

We now describe the various experiments conducted on our Marathi- Hindi and Hindi- Marathi Baseline SMT system[1] (Sreelekha et. al., 2013) by augmenting various lexical resources and the comparisons of results in the form of an error analysis. We have used Moses (Koehn et. al., 2007) and Giza++[2] for modeling the baseline system. Table 1 shows the statistics of corpus and the various lexical resources used for our experiments. Various experiments are conducted on top of the baseline system, such as; with an uncleaned corpus, with a cleaned corpus, with IndoWordnet extracted words, with Suffix splitted corpus, with Function words and Kridantha pairs and with verb phrases. The statistics of lexical resources used are shown in Table 1 and the results are shown in Tables 2, 3, 4 and 5. The lexical resources were extracted programmatically and have been validated manually with two Marathi-Hindi bilingual experts, with qualifications of Master degree in Hindi and Marathi Literature, over a period of 2 years. The detailed description of each experiment is explained with an example as listed below:

### 3.1 Baseline system with an unclean corpus

The corpus used for our experiments are taken from ILCI and DIT corpus in Tourism and Health domain. The corpus was contaminated with stylistic constructions which prevent the learning of correct grammatical structures, mis-alignments, wrong and missing translations which affected the quality of translation. Consider a sentence from the uncleaned Marathi-Hindi corpus, where the translation is wrong,

**Marathi** : जेवणात जास्त मिरची-मसाले व आम्लीय रसांपासून बनवलेल्या खाद्यपदार्थांचे सेवन केल्याने रीरात आम्लता जास्त वाढते. *{jevanaat jaast mirchi-masale va amleey rasampasoon banvalelya khachpa darthache sevan kelyane sareerat aamlata jaast vadte} {Because of the consumption of food made with more chilli - spices and acidic juices acidity increases more in the body.}*

**Equivalent Hindi Translation** (wrong) : आहार मिर्च- मसाले और अम्लीय संतुलन से बना भोजन की खपत की तुलना में अधिक शरीर में अम्लता बढ़ जाती है। *{aahaar mirch - masaale aur amleey santulan se bana bhojan kee khapat kee tulana mein adhik shareer mein amlata badh jaatee hai.} {Food chili - made from spices and food consumption more than the acidic balance in the body increases the acidity .}*

Here, the above Hindi translation is wrong and the statistical models generated by training with the uncleaned corpus also results in poor quality translation. The experiemntal results with uncleaned corpus is shown in the Table 2, 3, 4 and 5. During the error analysis we observed that the quality of the the parallel corpus plays a major role in generating the good quality translation models. Hence we focussed on cleaning the parallel corpus before training.

### 3.2 Baseline system with cleaned corpus

The Marathi-Hindi bilingual experts have cleaned the parallel corpus such as, removed the stylistic constructions, unwanted characters and wrong translations and corrected the missing translations and phrases. In order to improve the word-word alignments learning, we have manually aligned the source and target sentences in the parallel corpus.

Consider the above discussed wrongly translated Marathi sentence, **Marathi :** जेवणात जास्त मिरची-मसाले व आम्लीय रसांपासून बनवलेल्या खाद्यपदार्थांचे सेवन केल्याने शरीरात आम्लता जास्त वाढते . *{jevanaat jaast mirchi-masale va amleey rasampasoon banvalelya khachpa darthache sevan kelyane sareerat aamlata jaast vadte} {Because of the consumption of food made with more chilli - spices and acidic juices acidity increases more in the body.}*

After the cleaning process, the Marathi sentence has been correctly translated into Hindi as,

**Correct Hindi Translation :** भोजन में अधिक मिर्च-मसालों व अम्लीय रसों से बने खाद्य पदार्थों का सेवन करने से शरीर में अम्लता अधिक बढ़ती है । *{bhojan men adhik mirch-masalom va amleeya rasom se bane khaach padarthon ka sevan karne se sareer men amlatha adhik badthi hae} {Because of the consumption of food made with more chilli - spices and acidic juices acidity increases more in the body.}*

After training with the cleaned corpus, MT system was able to generate good quality translations. As shown in Table 2, 3, 4 and 5, the quality of the translation has improved to more than 40% compared to SMT system with uncleanted corpus. After a detailed error analysis we observed that the system fails to handle the rich morphology and we started investigation to handle the morphological inflections.

### 3.3 SMT system with Suffix splitted corpus

In order to handle the rich morphology we conducted experimenting with suffix splitting of the agglutinated inflected words. Consider a Marathi sentence,

**Marathi :** ही सात धर्मस्थळे सात नगरी वा सप्तपुरींच्या रूपात ग्रंथांमध्ये वर्णिलेली आहेत. *{hi saat dharmstale saat nagari va saptapureechya roopant grandhamadhye varnaleli aahet} {These seven shrine seven towns are described in the texts as Sptpurion .}* **Marathi sentence with suffix split:** ही सात धर्मस्थळे सात नगर ई वा सप्तपुरी च्या रूप त ग्रंथ मध्ये वर्ण लेली अस. *{hi saat dharmstale saat nagar yi va saptapuri chya roop th grandh madhye varn leli asu}*

We have done splitting of agglutinative suffixes for the inflected words in the entire corpus and after that the experiments were conducted. After analyzing the results of suffix splitted corpus trained models, we have observed that even though agglutinative suffixes are getting splitted, it increases alignment options and hence the quality of the translation is not improving in a great level. As shown in Table 2, 3, 4 and 5, the quality of the translation has improved slightly more than Baseline SMT system. After a detailed error analysis we decided to experiment with IndoWordnet synset words to handle the vocabulary differences and ambiguity.

---
[1] http://www.cfilt.iitb.ac.in/SMTSystem
[2] http://www.statmt.org/

| Sl. No | Corpus Source | Training Corpus [Manually cleaned and aligned] | Corpus Size [Sentences] |
|---|---|---|---|
| 1 | ILCI | Tourism | 24250 |
| 2 | ILCI | Health | 24250 |
| 3 | DIT | Tourism | 20000 |
| 4 | DIT | Health | 20000 |
| | | Total | 88500 |

| Sl. No | Lexical Resource Source | Lexical Resources in Corpus | Lexical Resource Size [Words] |
|---|---|---|---|
| 1 | CFILT, IIT Bombay | IndoWordnet Synset words | 450000 |
| 2 | CFILT IIT Bombay | Function Words, Kridanata Pairs | 5000 |
| 3 | CFILT IIT B | Verb Phrases | 15000 |
| | | Total | 470000 |

| Sl. No | Corpus Source ILCI | Tuning Corpus Size | Testing Corpus Size [Sentences] |
|---|---|---|---|
| 1 | Tourism, Health | 500 | 1000 |

Table 1: Statistics of Corpus and Lexical Resources Used

### 3.4 With IndoWordnet extracted words

The bilingually mapped words of 450000 were extracted from Indowordnet [16] with its semantic and lexical relations. We have considered all the possible synset word mappings for a single word and generated that many entries of parallel words. Consider the word गुस्सा_करना {gussa-karna}{getting angry} and it's generated synset wordmappings from IndoWordnet.

गुस्सा_करना : चिडणे संतापणे भडकणे कोपणे चिरडणे {gussa karna: chidne santhapane bhadkane kopane chiradane } {getting-angry:little-angry moderate-angry very-much-angry getting-angry getting-angry }

The extracted Indowordnet synset words were augmented into the training corpus and the results compared with the baseline system are shown in the Table 2, 3, 4 and 5. We have observed the augmenting with synset words not only helped in improving the quality of translation but also it helped in handing the lexical and semantic ambiguity as well. Also we analyzed that the resultant translation fails to handle various inflected forms, kridanta forms, case markers etcetera at various times. So we further investigated on handling this issue and decided to construct a parallel function words and kridanta pairs.

### 3.5 With Function words and Kridantha pairs

During our case study we observed that Marathi and Hindi takes 7 types of kridanta forms along with its' post position, pre-position and inflected forms. Hence we have prepared 5000 parallel entries of kridanta, akhyat, function words, suffix pairs etc over a period of 2 months and augmented it into the training corpus.

Consider a sample Marathi-Hindi kridanta form pair,
खाकर : खाऊन { khakar : khaoon} { ate : ate}

Upon analyzing the resultant translations we observed that grammatical structure as well as the quality of the translation has improved a lot. Comparative results are shown in the Table 2, 3, 4 and 5. During the error analysis we observed that even though quality and structure of the translation is improving; it fails to handle the verbal inflections properly. Further we started a study on verbal inflections since Marathi is having rich morphology with highly agglutinative suffixes.

### 3.6 Corpus with verb phrases

We have prepared and validated 15000 entries of Marathi-Hindi verb phrases over a period of 3 months, which contain plentiful examples of various verbal inflections and augmented it into the training corpus.

Consider a sample verb phrase entry from the training corpus, दर्शन के समय : दर्शनाच्या वेळी {darshan ke samay: darshanachya veli} {visting time : visiting time}

We have observed that the MT system was able to translate the verb phrases correctly to a great extent. The error analysis study shows that the quality of the translation has improved drastically as shown in the Tables 2, 3, 4 and 5.

## 4. Evaluation & Error Analysis

| Hindi-Marathi SMT System | | BLEU score | METEOR | TER |
|---|---|---|---|---|
| Baseline system with Uncleaned Corpus | Without Tuning | 2.06 | 0.123 | 95.48 |
| | With Tuning | 2.76 | 0.124 | 93.94 |
| Baseline system with Cleaned Corpus | Without Tuning | 23.97 | 0.190 | 65.05 |
| | With Tuning | 27.76 | 0.193 | 63.92 |
| Baseline system with Suffix Split Corpus | Without Tuning | 24.97 | 0.191 | 64.05 |
| | With Tuning | 28.76 | 0.194 | 63.12 |
| Corpus with Wordnet | Without Tuning | 29.31 | 0.243 | 54.91 |
| | With Tuning | 31.78 | 0.257 | 52.30 |
| Corpus with FunctionWords, kridanta pairs | Without Tuning | 33.25 | 0.274 | 46.79 |
| | With Tuning | 35.21 | 0.282 | 42.06 |
| Corpus With Verb Phrases | Without Tuning | 39.26 | 0.301 | 35.06 |
| | With Tuning | 42.15 | 0.321 | 31.55 |

Table 2: Results of Hindi-Marathi SMT BLEU score, METEOR, TER Evaluations

We have tested the translation system with a corpus of 2000 sentences taken from the 'ILCI tourism, health' corpus as shown in Table 1. In addition we have used a tuning (MERT) corpus of 500 sentences as shown in Table 1. We have evaluated the translated outputs of both Marathi to Hindi and Hindi to Marathi SMT systems in all 5 categories using various methods such as subjective evaluation, BLEU score (Papineni et al., 2002), METEOR and TER (Agarwal and Lavie 2008). We gave importance to subjective evaluation to determine the fluency (F) and adequacy (A) of the translation, since for morphologically rich languages subjective evaluations can give more accurate results compared to others. We have followed the subjective evaluation procedure as described in Sreelekha et.al.(2013) and the results are given in Table 4 and 5. The results of BLEU, METEOR and TER evaluations are displayed in Tables 2 and 3.

| Marathi-Hindi SMT System | | BLEU score | METEOR | TER |
|---|---|---|---|---|
| Baseline system with Uncleaned Corpus | Without Tuning | 2.16 | 0.121 | 90.32 |
| | With Tuning | 2.80 | 0.125 | 89.52 |
| Baseline system with Cleaned Corpus | Without Tuning | 24.01 | 0.181 | 68.32 |
| | With Tuning | 26.82 | 0.186 | 67.89 |
| Baseline system with Suffix Split Corpus | Without Tuning | 27.01 | 0.183 | 66.32 |
| | With Tuning | 28.22 | 0.188 | 65.89 |
| Corpus with Wordnet | Without Tuning | 30.54 | 0.278 | 56.75 |
| | With Tuning | 34.48 | 0.281 | 54.30 |
| Corpus-with FunctionWords, kridanta pairs | Without Tuning | 38.20 | 0.286 | 48.19 |
| | With Tuning | 41.46 | 0.293 | 46.19 |
| Corpus with Verb Phrases | Without Tuning | 48.80 | 0.323 | 39.36 |
| | With Tuning | 51.60 | 0.334 | 35.48 |

**Table 3: Results of Marathi-Hindi SMT BLEU score, METEOR, NER Evaluations**

| Marathi-Hindi SMT System | | Adequacy | Fluency |
|---|---|---|---|
| Baseline system with an uncleaned corpus | Without Tuning | 17.8% | 22.34% |
| | With Tuning | 20.6% | 30.8% |
| Baseline system with Cleaned Corpus | Without Tuning | 55.76% | 66.12% |
| | With Tuning | 60.6% | 71.3% |
| Baseline system with Suffix Split | Without Tuning | 59.18% | 67.67% |
| | With Tuning | 61.76% | 72.31% |
| Corpus with Wordnet | Without Tuning | 71.6% | 80.2% |
| | With Tuning | 74.76% | 83.89% |
| Corpus with Function Words, kridanta pairs | Without Tuning | 78% | 86.87% |
| | With Tuning | 80.81% | 88.12% |
| Corpus With Verb Phrases | Without Tuning | 84.34% | 87.78% |
| | With Tuning | 87.67% | 90.35% |

**Table 4 : Marathi-Hindi SMT Subjective Evaluation Results**

| Hindi-Marathi Statistical MT System | | Adequacy | Fluency |
|---|---|---|---|
| Baseline system with Uncleaned Corpus | Without Tuning | 15.56% | 21.67% |
| | With Tuning | 19.98% | 27.39% |
| Baseline system with Cleaned Corpus | Without Tuning | 54.98% | 65.67% |
| | With Tuning | 59.16% | 71.21% |
| Baseline system with Suffix Split | Without Tuning | 55.52% | 66.97% |
| | With Tuning | 59.76% | 72.11% |
| Corpus with Wordnet | Without Tuning | 69.09% | 79.14% |
| | With Tuning | 72.73% | 82.65% |
| Corpus with Function Words, kridanta pairs | Without Tuning | 76.13% | 85.68% |
| | With Tuning | 79.36% | 87.21% |
| Corpus With Verb Phrases | Without Tuning | 82.65% | 86.34% |
| | With Tuning | 86.01% | 89.32% |

**Table 5: Hindi-Marathi SMT System Subjective Evaluation Results**

We have observed that the quality of the translation is improving as the corpus is getting cleaned and more lexical resources are being used. Hence, there is an incremental growth in adequacy, fluency, BLEU, METEOR scores and reduction in TER score. The performance comparison graph is shown in figure 1 and figure 2. The fluency of the translation is increased up to 90.35% in the case of Marathi to Hindi and up to 89.32% in the case of Hindi to Marathi, which is 4 times more than the baseline system results.

**Figure 1: Hindi-Marathi SMT Evaluation Analysis**

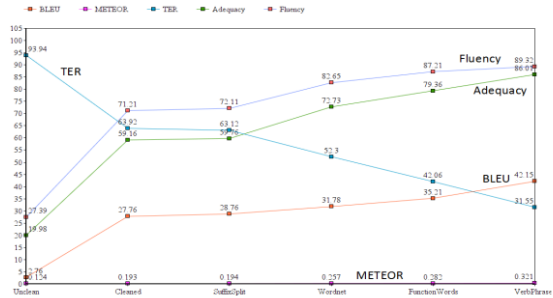

**Figure 2 : Marathi-Hindi SMT Evaluation Analysis**

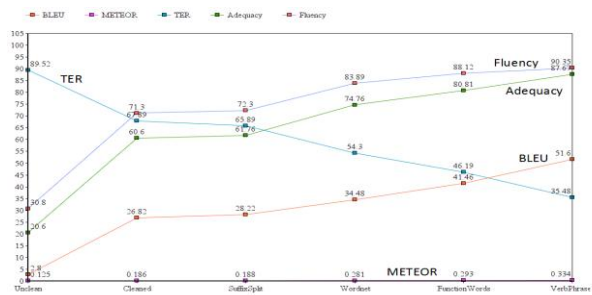

## 5. Conclusion

In this paper, we have investigated on various ways to improve the quality of machine translation in two resource poor language Marathi and Hindi. We have prepared and experimented with various lexical resources such as lexical words, function words, kridanta pairs and verb phrases etcetera. We have discussed the six categories of experiments on top of the baseline phrase based SMT system with 24 trained models and its comparative performance in detail for both Marathi–Hindi and Hindi-Marathi pairs. With the help of augmented lexical resources, SMT system was able to handle the morphological infections and grammatical structures to a great extend. In order to do a thorough evaluation, we have used various measures such as BLEU Score, METEOR, TER, subjective evaluations in terms of Fluency and Adequacy. Evaluation results show that there is an incremental growth for both Marathi-Hindi and Hindi-Marathi systems in terms of BLEU-Score, METEOR, Adequacy and Fluency. There is a gradual reduction in TER evaluation scores, which shows the improvement in translation quality. These experiments give an insight into the utilization of various lexical resources for an efficient Machine Translation system development for resource poor languages. Our future work will be focused on investigating more lexical resources for improving the quality of Statistical Machine Translation systems for various language pairs.

**Acknowledgments**

This work is funded by Department of Science and Technology, Govt. of India under Women Scientist Scheme- WOS-A with the project code- SR/WOS-A/ET-1075/2014.